%% file: pgds.tex
\documentclass{article}

\usepackage[final]{nips_2016}
\PassOptionsToPackage{numbers, compress}{natbib}

\usepackage[utf8]{inputenc} % allow utf-8 input
\usepackage[T1]{fontenc}    % use 8-bit T1 fonts
\usepackage{hyperref}       % hyperlinks
\usepackage{url}            % simple URL typesetting
\usepackage{booktabs}       % professional-quality tables
\usepackage{amsfonts}       % blackboard math symbols
\usepackage{nicefrac}       % compact symbols for 1/2, etc.
\usepackage{microtype}      % microtypography

\usepackage{amsmath,relsize,mdframed,wrapfig}
\usepackage{setspace}
\usepackage[pdftex]{graphicx}

\usepackage[plain]{algorithm}
\usepackage[noend]{algpseudocode}

\usepackage{dsfont}

\input{macros}

\def\everything1{\mathrel{\ooalign{%
        \raisebox{0.5\height}{$\scriptstyle 1, \ldots, T$}\cr\hidewidth\raisebox{-0.6\height}{$-\,$}\hidewidth\cr}}}

\title{Poisson--Gamma Dynamical Systems}

\author{
  Aaron Schein \\
  College of Information and Computer Sciences\\
  University of Massachusetts Amherst\\
  Amherst, MA 01003 \\
  \texttt{aschein@cs.umass.edu}
  \AND
  Mingyuan Zhou \\
  McCombs School of Business \\
  The University of Texas at Austin \\
  Austin, TX 78712 \\
  \texttt{mingyuan.zhou@mccombs.utexas.edu}
  \And
  Hanna Wallach \\
  Microsoft Research New York \\
  641 Avenue of the Americas \\
  New York, NY 10011 \\
  \texttt{hanna@dirichlet.net}
}

\begin{document}

\maketitle

%% TODO: think about augmentations. that relies on recently dveloped
%% theory.

\begin{abstract}
We introduce a new dynamical system for sequentially observed
multivariate count data. This model is based on the gamma--Poisson
construction---a natural choice for count data---and relies on a novel
Bayesian nonparametric prior that ties and shrinks the model
parameters, thus avoiding overfitting. We present an efficient MCMC
inference algorithm that advances recent work on augmentation schemes
for inference in negative binomial models. Finally, we demonstrate the
model's inductive bias using a variety of real-world data sets,
showing that it exhibits superior predictive performance over other
models and infers highly interpretable latent structure.~\looseness=-1
\end{abstract}

\section{Introduction}
\label{sec:introduction}

Sequentially observed count vectors $\boldsymbol{y}^{(1)}, \ldots,
\boldsymbol{y}^{(T)}$ are the main object of study in many real-world
applications, including text analysis, social network analysis, and
recommender systems. Count data pose unique statistical and
computational challenges when they are high-dimensional, sparse, and
overdispersed, as is often the case in real-world applications. For
example, when tracking counts of user interactions in a social
network, only a tiny fraction of possible edges are ever active,
exhibiting bursty periods of activity when they are. Models of such
data should exploit this sparsity in order to scale to high dimensions
and be robust to overdispersed temporal patterns. In addition to these
characteristics, sequentially observed multivariate count data often
exhibit complex dependencies within and across time steps. For
example, scientific papers about one topic may encourage researchers
to write papers about another related topic in the following
year. Models of such data should therefore capture the topic structure
of individual documents as well as the excitatory relationships
between topics.\looseness=-1

The linear dynamical system (LDS) is a widely used model for
sequentially observed data, with many well-developed inference
techniques based on the Kalman
filter~\cite{kalman1960new,ghahramani1999learning}. The LDS assumes
that each sequentially observed $V$-dimensional vector
$\boldsymbol{r}^{(t)}$ is real valued and Gaussian distributed:
$\boldsymbol{r}^{(t)} \sim
\mathcal{N}(\Phi\,\boldsymbol{\theta}^{(t)}, \Sigma)$, where
$\boldsymbol{\theta}^{(t)} \in \mathbb{R}^K$ is a latent state, with
$K$ components, that is linked to the observed space via $\Phi \in
\mathbb{R}^{V\times K}$. The LDS derives its expressive power from the
way it assumes that the latent states evolve:
$\boldsymbol{\theta}^{(t)} \sim \mathcal{N}(\Pi\,
\boldsymbol{\theta}^{(t-1)}, \Delta)$, where $\Pi \in \mathbb{R}^{K
  \times K}$ is a transition matrix that captures between-component
dependencies across time steps. Although the LDS can be linked to
non-real observations via the extended Kalman
filter~\cite{haykin2001kalman}, it cannot efficiently model real-world
count data because inference is $\mathcal{O}((K + V)^3)$ and thus
scales poorly with the dimensionality of the
data~\cite{ghahramani1999learning}.\looseness=-1

Many previous approaches to modeling sequentially observed count data
rely on the generalized linear modeling
framework~\cite{mccullagh1989generalized} to link the observations to
a latent Gaussian space---e.g., via the Poisson--lognormal
link~\cite{bulmer1974fitting}. Researchers have used this construction
to factorize sequentially observed count matrices under a Poisson
likelihood, while modeling the temporal structure using well-studied
Gaussian techniques~\cite{blei2006dynamic,charlin2015dynamic}. Most of
these previous approaches assume a simple Gaussian state-space
model---i.e., $\boldsymbol{\theta}^{(t)} \sim
\mathcal{N}(\boldsymbol{\theta}^{(t-1)}, \Delta)$---that lacks the
expressive transition structure of the LDS; one notable exception is the Poisson linear dynamical system~\cite{macke2011empirical}. In practice, these approaches
exhibit prohibitive computational complexity in high dimensions, and
the Gaussian assumption may fail to accommodate the burstiness often
inherent to real-world count data~\cite{kleinberg2003bursty}.
\begin{wrapfigure}{r}{6cm}
\centering
\vspace{2.9mm}
\includegraphics[width=0.95\linewidth]{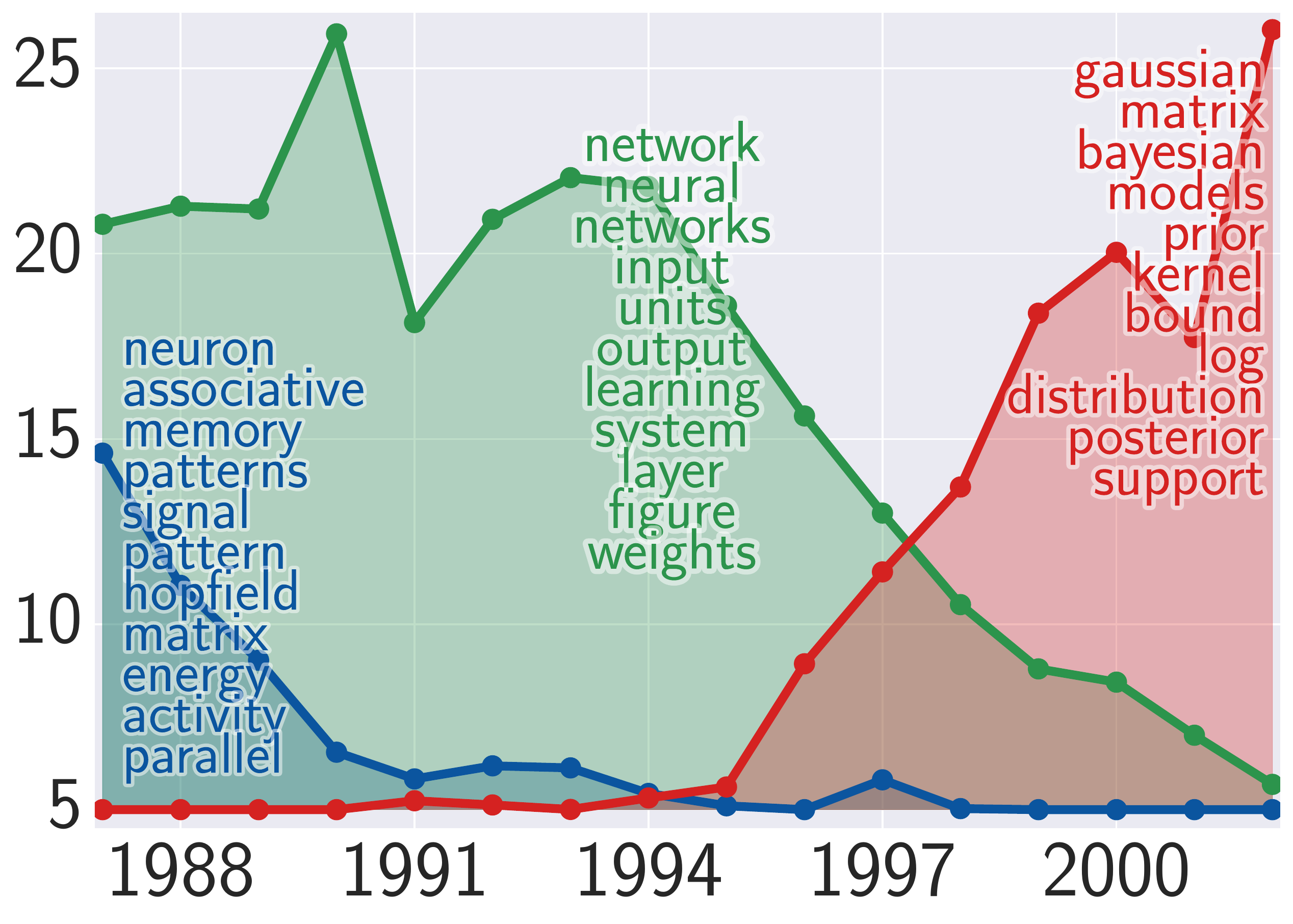}
%\vspace{-4mm}
\caption{\label{fig:nips} The time-step factors for three components
  inferred by the PGDS from a corpus of NIPS papers. Each component is
  associated with a feature factor for each word type in the corpus;
  we list the words with the largest factors.  The inferred structure
  tells a familiar story about the rise and fall of certain subfields
  of machine learning.}
\vspace{-7mm}
\end{wrapfigure}

We present the \emph{Poisson--gamma dynamical system (PGDS)}---a new
dynamical system, based on the gamma--Poisson construction, that
supports the expressive transition structure of the LDS. This model
naturally handles overdispersed data. We introduce a new Bayesian
nonparametric prior to automatically infer the model's rank. We
develop an elegant and efficient algorithm for inferring the
parameters of the transition structure that advances recent
work on augmentation schemes for inference in negative binomial
models~\cite{zhou12augment-and-conquer} and scales with the number of non-zero counts, thus exploiting the sparsity inherent to real-world count data. We examine the way in which the dynamical
gamma--Poisson construction propagates information and derive the
model's steady state, which involves the Lambert W
function~\cite{corless1996lambertw}. Finally, we use the PGDS to
analyze a diverse range of real-world data sets, showing that it
exhibits excellent predictive performance on smoothing and forecasting
tasks and infers interpretable latent structure, an example of which
is depicted in figure~\ref{fig:nips}.

\section{Poisson--Gamma Dynamical Systems}
\label{sec:model}

We can represent a data set of $V$-dimensional sequentially observed
count vectors $\boldsymbol{y}^{(1)}, \ldots, \boldsymbol{y}^{(T)}$ as
a $V \times T$ count matrix $Y$. The PGDS models a single count
$y_v^{(t)} \in \{0, 1, \ldots\}$ in this matrix as
follows:\looseness=-1
\begin{equation}
  \label{eqn:model}
  y_v^{(t)} \sim \textrm{Pois}(\delta^{(t)}\,\textstyle{\sum}_{k=1}^K
  \phi_{vk}\,\theta_k^{(t)})\textrm{ and\ \,} \theta^{(t)}_k \sim
  \textrm{Gam}(\tau_0\, \textstyle{\sum}_{k_2=1}^K \pi_{kk_2}\,
  \theta_{k_2}^{(t-1)}, \tau_0),
\end{equation}
where the latent factors $\phi_{vk}$ and $\theta_k^{(t)}$ are both
positive, and represent the strength of feature $v$ in component $k$
and the strength of component $k$ at time step $t$, respectively. The
scaling factor $\delta^{(t)}$ captures the scale of the counts at time
step $t$, and therefore obviates the need to rescale the data as a
preprocessing step. We refer to the PGDS as \emph{stationary} if
$\delta^{(t)} \teq \delta$ for $t = 1, \ldots, T$. We can view the
feature factors as a $V \times K$ matrix $\Phi$ and the time-step
factors as a $T \times K$ matrix $\Theta$. Because we can also
collectively view the scaling factors and time-step factors as a $T
\times K$ matrix $\Psi$, where element $\psi_{tk} =
\delta^{(t)}\,\theta_k^{(t)}$, the PGDS is a form of Poisson matrix
factorization: $Y \sim
\textrm{Pois}(\Phi\,\Psi^T)$~\citep{canny2004gap,
  cemgil09bayesian,zhou2011beta,gopalan15scalable}.

The PGDS is characterized by its expressive transition structure,
which assumes that each time-step factor $\theta_k^{(t)}$ is drawn
from a gamma distribution, whose shape parameter is a linear
combination of the $K$ factors at the previous time step. The latent
transition weights $\pi_{11}, \ldots, \pi_{k_1 k_2}, \ldots,
\pi_{KK}$, which we can view as a $K \times K$ transition matrix
$\Pi$, capture the excitatory relationships between components. The
vector $\boldsymbol{\theta}^{(t)} = (\theta_1^{(t)}, \ldots,
\theta_K^{(t)})$ has an expected value of
$\mathbb{E}[\boldsymbol{\theta}^{(t)} \g \boldsymbol{\theta}^{(t-1)},
  \Pi] = \Pi\, \boldsymbol{\theta}^{(t-1)}$ and is therefore analogous
to a latent state in the the LDS. The concentration parameter $\tau_0$
determines the variance of $\boldsymbol{\theta}^{(t)}$---specifically,
$\textrm{Var}\,(\boldsymbol{\theta}^{(t)} \g
\boldsymbol{\theta}^{(t-1)}, \Pi) =
(\Pi\,\boldsymbol{\theta}^{(t-1)})\, \tau_0^{-1}$---without affecting
its expected value.\looseness=-1

To model the strength of each component, we introduce $K$ component
weights $\boldsymbol{\nu} = (\nu_1,\dots,\nu_K)$ and place a shrinkage
prior over them. We assume that the time-step factors and transition
weights for component $k$ are tied to its component weight
$\nu_k$. Specifically, we define the following structure:\looseness=-1
\begin{equation}
  \label{eqn:priors}
  \theta_k^{(1)} \sim \textrm{Gam}(\tau_0\,\nu_k, \tau_0)\textrm{
    and\ \,} \boldsymbol{\pi}_k \sim \textrm{Dir}(\nu_1\nu_k, \ldots,
  \xi\nu_k\, \ldots, \nu_K\nu_k) \textrm{ and\ \,} \nu_k \sim
  \textrm{Gam}(\tfrac{\gamma_0}{K}, \beta),
\end{equation}
where $\boldsymbol{\pi}_k = (\pi_{1k}, \ldots, \pi_{Kk})$ is the
$k^{\textrm{th}}$ column of $\Pi$. Because $\sum_{k_1=1}^K \pi_{k_1 k} = 1$, we can interpret
$\pi_{k_1 k}$ as the probability of transitioning from component $k$
to component $k_1$. (We note that interpreting $\Pi$ as a stochastic
transition matrix relates the PGDS to the discrete hidden Markov
model.) For a fixed value of $\gamma_0$,
increasing $K$ will encourage many of the component weights to be
small. A small value of $\nu_k$ will shrink $\theta^{(1)}_k$, as well
as the transition weights in the $k^{\textrm{th}}$ row of
$\Pi$. Small values of the transition weights in the
$k^{\textrm{th}}$ row of $\Pi$ therefore prevent component $k$ from
being excited by the other components and by itself. Specifically,
because the shape parameter for the gamma prior over $\theta^{(t)}_k$
involves a linear combination of $\boldsymbol{\theta}^{(t-1)}$ and the
transition weights in the $k^{\textrm{th}}$ row of $\Pi$, small
transition weights will result in a small shape parameter, shrinking
$\theta^{(t)}_k$. Thus, the component weights play a critical role in
the PGDS by enabling it to automatically turn off any unneeded
capacity and avoid overfitting.

Finally, we place Dirichlet priors over the feature factors and draw
the other parameters from a non-informative gamma prior:
$\boldsymbol{\phi}_k = (\phi_{1k}, \ldots, \phi_{Vk}) \sim
\textrm{Dir}(\eta_0, \ldots, \eta_0)$ and $\delta^{(t)}, \xi, \beta
\sim \textrm{Gam}(\epsilon_0, \epsilon_0)$. The PGDS therefore has
four positive hyperparameters to be set by the user: $\tau_0$,
$\gamma_0$, $\eta_0$, and $\epsilon_0$.

\textbf{Bayesian nonparametric interpretation:} As $K \rightarrow
\infty$, the component weights and their corresponding feature factor
vectors constitute a draw $G = \sum_{k=1}^\infty \nu_k
\mathds{1}_{\boldsymbol{\phi}_k}$ from a gamma process
$\textrm{GamP}\left(G_0,\, \beta\right)$, where $\beta$ is a scale
parameter and $G_0$ is a finite and continuous base measure over a
complete separable metric space
$\Omega$~\cite{ferguson73bayesian}. Models based on the gamma process
have an inherent shrinkage mechanism because the number of atoms with
weights greater than $\varepsilon > 0$ follows a Poisson distribution
with a finite mean---specifically, $\textrm{Pois}(\gamma_0
\int_{\varepsilon}^\infty \textrm{d}\nu\ \nu^{-1}
\exp{(-\beta\,\nu)})$, where $\gamma_0 = G_0(\Omega)$ is the total
mass under the base measure. This interpretation enables us to view
the priors over $\Pi$ and $\Theta$ as novel stochastic processes,
which we call the \emph{column-normalized relational gamma process}
and the \emph{recurrent gamma process}, respectively. We provide the
definitions of these processes in the supplementary material.

\textbf{Non-count observations:} The PGDS can also model non-count
data by linking the observed vectors to latent counts. A binary
observation $b_v^{(t)}$ can be linked to a latent Poisson count
$y_v^{(t)}$ via the Bernoulli--Poisson distribution: $b_{v}^{(t)} =
\mathds{1}(y_{v}^{(t)}\geq 1)$ and $y_{v}^{(t)} \sim
\textrm{Pois}(\delta^{(t)}\sum_{k=1}^K \phi_{vk}\,
\theta^{(t)}_k)$~\citep{zhou15infinite}. Similarly, a real-valued
observation $r_v^{(t)}$ can be linked to a latent Poisson count
$y_v^{(t)}$ via the Poisson randomized gamma
distribution~\cite{zhou2015gamma}. Finally, Basbug and
Engelhardt~\citep{basbug2016hierarchical} recently showed that many
types of non-count matrices can be linked to a latent count matrix via
the compound Poisson distribution~\citep{adelson1966compound}.

\section{MCMC Inference}
\label{sec:inference}   

MCMC inference for the PGDS consists of drawing samples of the model
parameters from their joint posterior distribution given an observed
count matrix $Y$ and the model hyperparameters $\tau_0$, $\gamma_0$,
$\eta_0$, $\epsilon_0$. In this section, we present a Gibbs sampling
algorithm for drawing these samples. At a high level, our approach is
similar to that used to develop Gibbs sampling algorithms for several
other related
models~\cite{zhou12augment-and-conquer,zhou15negative,acharya2015nonparametric,zhou15infinite};
however, we extend this approach to handle the unique properties of
the PGDS. The main technical challenge is sampling $\Theta$ from its
conditional posterior, which does not have a closed form. We address
this challenge by introducing a set of auxiliary variables. Under this
augmented version of the model, marginalizing over $\Theta$ becomes
tractable and its conditional posterior has a closed form. Moreover,
by introducing these auxiliary variables and marginalizing over
$\Theta$, we obtain an alternative model specification that we can
subsequently exploit to obtain closed-form conditional posteriors for
$\Pi$, $\boldsymbol{\nu}$, and $\xi$. We marginalize over $\Theta$ by
performing a ``backward filtering'' pass, starting with
$\boldsymbol{\theta}^{(T)}$. We repeatedly exploit the following three
definitions in order to do this.

% Do not start a sentence with a mathematical variable.
% We don't use ind in this context elsewhere in the paper, so for
% consistency we should not use it here. We should also follow the
% sentence structure of the other definitions.

% We're using y and theta for a general variables here, which is not
% consistent with our use of them as a specific variables elsewhere.

\textit{Definition 1:} If $y_{\boldsymbol{\cdot}} \!=\! \sum_{n=1}^N y_n$,
where $y_n \sim \textrm{Pois}(\theta_n)$ are independent
Poisson-distributed random variables, then $(y_1, \ldots, y_N) \sim
\textrm{Mult}(y_{\boldsymbol{\cdot}}, (\frac{\theta_1}{\sum_{n=1}^N
  \theta_n}, \ldots, \frac{\theta_N}{\sum_{n=1}^N \theta_n}))$ and
$y_{\boldsymbol{\cdot}} \sim \textrm{Pois}(\sum_{n=1}^N
\theta_n)$~\cite{kingman72poisson,Dunson05bayesianlatent}.\looseness=-1

\textit{Definition 2:} If $y \sim \textrm{Pois}(c\,\theta)$, where $c$
is a constant, and $\theta \sim \textrm{Gam}(a, b)$, then $y \sim
\textrm{NB}(a, \frac{c}{b + c})$ is a negative binomial--distributed
random variable. We can equivalently parameterize it as $y \sim
\textrm{NB}(a, g(\zeta))$, where $g(z) = 1 - \exp{(-z)}$ is the
Bernoulli--Poisson link~\cite{zhou15infinite} and $\zeta = \ln{(1 +
  \frac{c}{b})}$. \looseness=-1

% u usually means a uniform random numbers, so let's not use it here.

\textit{Definition 3:} If $y \sim \textrm{NB}(a, g(\zeta))$ and $l
\sim \textrm{CRT}(y, a)$ is a Chinese restaurant table--distributed
random variable, then $y$ and $l$ are equivalently jointly distributed
as $y \sim \textrm{SumLog}(l, g(\zeta))$ and $l \sim \textrm{Pois}(a\,
\zeta)$~\cite{zhou15negative}. The sum logarithmic distribution is
further defined as the sum of $l$ independent and identically
logarithmic-distributed random variables---i.e., $y = \sum_{i=1}^l
x_i$ and $x_i \sim \textrm{Log}(g(\zeta))$.

\textbf{Marginalizing over $\Theta$:} We first note that we can
re-express the Poisson likelihood in equation~\ref{eqn:model} in terms
of latent subcounts~\cite{cemgil09bayesian}: $y_{v}^{(t)} = y_{v
  \boldsymbol{\cdot}}^{(t)} = \sum_{k=1}^K y_{v k}^{(t)}$ and
$y_{vk}^{(t)} \sim \Pois(\delta^{(t)}\,\phi_{vk}\,\theta_k^{(t)})$. We
then define $y_{\boldsymbol{\cdot} k}^{(t)} = \sum_{v=1}^V
y_{vk}^{(t)}$. Via definition 1, we obtain $y_{\boldsymbol{\cdot}
  k}^{(t)} \sim \textrm{Pois}(\delta^{(t)}\,\theta_k^{(t)})$ because
$\sum_{v=1}^V \phi_{vk} = 1$.\looseness=-1

We start with $\theta_k^{(T)}$ because none of the other time-step
factors depend on it in their priors. Via definition 2, we
can immediately marginalize over $\theta_k^{(T)}$ to obtain the following equation:
\begin{equation}
  \label{eqn:y_T}
y_{\boldsymbol{\cdot} k}^{(T)} \sim \textrm{NB}(\tau_0 \textstyle{\sum}_{k_2=1}^K \pi_{k
  k_2}\,\theta_{k_2}^{(T-1)}, g(\zeta^{(T)})) \textrm{, where\ \,}
\zeta^{(T)} = \ln{(1 + \frac{\delta^{(T)}}{\tau_0})}.
\end{equation}
Next, we marginalize over $\theta_k^{(T-1)}$. To do this, we introduce
an auxiliary variable: $l_{k}^{(T)} \sim
\textrm{CRT}(y_{\boldsymbol{\cdot} k}^{(T)}, \tau_0
\textstyle{\sum}_{k_2=1}^K \pi_{k k_2}\,\theta_{k_2}^{(T-1)})$. We can
then re-express the joint distribution over $y_{\boldsymbol{\cdot}
  k}^{(T)}$ and $l_{k}^{(T)}$ as\looseness=-1
\begin{equation}
\label{eqn:pois_sumlog}
  y_{\boldsymbol{\cdot} k}^{(T)} \sim \textrm{SumLog}(l_{k}^{(T)},
  g(\zeta^{(T)}) \textrm{ and\ \,} l_{k}^{(T)} \sim \textrm{Pois}(
  \zeta^{(T)}\, \tau_0 \textstyle{\sum}_{k_2=1}^K \pi_{k
    k_2}\,\theta_{k_2}^{(T-1)}).
\end{equation}
We are still unable to marginalize over $\theta_k^{(T-1)}$ because it
appears in a sum in the parameter of the Poisson distribution over
$l_{k}^{(T)}$; however, via definition 1, we can re-express this
distribution as
\begin{equation}
  l_k^{(T)} = l_{k \boldsymbol{\cdot}}^{(T)} = \textstyle{\sum}_{k_2=1}^K l_{k
    k_2}^{(T)} \textrm{ and\ \,}l_{k k_2}^{(T)} \sim
  \textrm{Pois}(\zeta^{(T)}\,\tau_0\,\pi_{k k_2}\,\theta_{k_2}^{(T-1)}).
\end{equation}
We then define $l_{\boldsymbol{\cdot} k}^{(T)} = \sum_{k_1=1}^K l_{k_1
  k}^{(T)}$. Again via definition 1, we can express the distribution over $l_{\boldsymbol{\cdot} k}^{(T)}$ as $l_{\boldsymbol{\cdot} k}^{(T)} \sim
\textrm{Pois}(\zeta^{(T)}\,\tau_0\,\theta_k^{(T-1)})$. We note that this expression
does not depend on the transition weights because $\sum_{k_1=1}^K
\pi_{k_1 k} = 1$. We also note that definition 1 implies that
$(l_{1k}^{(T)}, \ldots, l_{Kk}^{(T)}) \sim
\textrm{Mult}(l_{\boldsymbol{\cdot} k}^{(T)}, (\pi_1, \ldots,
\pi_K))$. Next, we introduce $m_k^{(T-1)} = y_{\boldsymbol{\cdot}
  k}^{(T-1)} + l_{\boldsymbol{\cdot} k}^{(T)}$, which summarizes all
of the information about the data at time steps $T-1$ and $T$ via
$y_{\boldsymbol{\cdot} k}^{(T-1)}$ and $l_{\boldsymbol{\cdot}
  k}^{(T)}$, respectively. Because $y_{\boldsymbol{\cdot} k}^{(T-1)}$
and $l_{\boldsymbol{\cdot} k}^{(T)}$ are both Poisson distributed, we
can use definition 1 to obtain
\begin{equation}
  \label{eqn:likelihood}
m_k^{(T-1)} \sim
\textrm{Pois}(\theta_k^{(T-1)}( \delta^{(T-1)} +
\zeta^{(T)}\,\tau_0)).
\end{equation}
Combining this likelihood with the gamma prior in
equation~\ref{eqn:model}, we can marginalize over
$\theta_k^{(T-1)}$:\looseness=-1
\begin{equation}
m_k^{(T-1)} \sim \textrm{NB}(\tau_0 \textstyle{\sum}_{k_2=1}^K \pi_{k
  k_2}\,\theta_{k_2}^{(T-2)}, g(\zeta^{(T-1)})) \textrm{, where\ \,}
\zeta^{(T-1)} = \ln{(1 + \frac{\delta^{(T-1)}}{\tau_0} + \zeta^{(T)})}.
\end{equation}
We then introduce $l_{k}^{(T-1)} \sim \textrm{CRT}(m_{k}^{(T-1)},
\tau_0 \sum_{k_2=1}^K \pi_{k k_2}\,\theta_{k_2}^{(T-2)})$ and
re-express the joint distribution over $l_{k}^{(T-1)}$ and
$m_k^{(T-1)}$ as the product of a Poisson and a sum logarithmic
distribution, similar to equation~\ref{eqn:pois_sumlog}. This then
allows us to marginalize over $\theta_k^{(T-2)}$ to obtain a negative
binomial distribution. We can repeat the same process all the way back
to $t=1$, where marginalizing over $\theta^{(1)}_k$ yields $m_k^{(1)}
\sim \textrm{NB}(\tau_0\,\nu_k, g(\zeta^{(1)}))$. We note that just as
$m_k^{(t)}$ summarizes all of the information about the data at time
steps $t, \ldots, T$, $\zeta^{(t)} = \ln{(1 +
  \frac{\delta^{(t)}}{\tau_0} + \zeta^{(t+1)})}$ summarizes all of the
information about $\delta^{(t)}, \ldots, \delta^{(T)}$.
\begin{wrapfigure}{r}{7.7cm}
  \centering
  \footnotesize
  \vspace{1.3mm}
\begin{algorithmic}
\State $l_{k \boldsymbol{\cdot}}^{(1)} \sim
\textrm{Pois}(\zeta^{(1)}\,\tau_0\,\nu_k)$
%\If{$t > 1$}
\State $(l_{1k}^{(t)}, \ldots, l_{Kk}^{(t)}) \sim
\textrm{Mult}(l_{\boldsymbol{\cdot} k}^{(t)}, (\pi_{1k}, \ldots,
\pi_{Kk}))$ for $t > 1$
\State $l_{k \boldsymbol{\cdot}}^{(t)} = \textstyle{\sum}_{k_2=1}^K
l_{k k_2}^{(t)}$ for $t > 1$
%\EndIf
\State $m_{k}^{(t)} \sim \textrm{SumLog}(l_{k \boldsymbol{\cdot}}^{(t)},
g(\zeta^{(t)}))$
\State $(y_{\boldsymbol{\cdot} k}^{(t)}, l_{\boldsymbol{\cdot} k}^{(t+1)}) \sim
\textrm{Bin}(m_k^{(t)}, (\frac{\delta^{(t)}}{\delta^{(t)} + \zeta^{(t+1)}\tau_0},\frac{\zeta^{(t+1)}\tau_0}{
  \delta^{(t)} + \zeta^{(t+1)}\tau_0}))$
\State $(y_{1k}^{(t)}, \ldots, y_{Vk}^{(t)}) \sim
\textrm{Mult}(y_{\boldsymbol{\cdot} k}^{(t)}, (\phi_{1k}, \ldots,
\phi_{Vk}))$
%\State $y_{v}^{(t)} = \sum_{k=1}^K y_{vk}^{(t)}$
\end{algorithmic}
\caption{\label{fig:alt_model}Alternative model specification.}
\vspace{-3em}
\end{wrapfigure}

As we mentioned previously, introducing these auxiliary variables and
marginalizing over $\Theta$ also enables us to define an alternative
model specification that we can exploit to obtain closed-form
conditional posteriors for $\Pi$, $\boldsymbol{\nu}$, and $\xi$. We
provide part of its generative process in
figure~\ref{fig:alt_model}. We define $m_k^{(T)} =
y_{\boldsymbol{\cdot} k}^{(T)} + l_{\boldsymbol{\cdot} k}^{(T+1)}$,
where $l_{\boldsymbol{\cdot} k}^{(T+1)} = 0$, and $\zeta^{(T+1)} = 0$
so that we can present the alternative model specification concisely.

\textbf{Steady state:} We draw particular attention to the backward
pass $\zeta^{(t)} = \ln{(1 + \frac{\delta^{(t)}}{\tau_0} +
  \zeta^{(t+1)})}$ that propagates information about
$\delta^{(t)},\dots,\delta^{(T)}$ as we marginalize over
$\Theta$. In the case of the stationary PGDS---i.e., $\delta^{(t)} =
\delta$---the backward pass has a fixed point that we define in the
following proposition.\looseness=-1

\textit{Proposition 1:} The backward pass has a fixed point of
$\zeta^{\star} = - \mathbb{W}_{-1}(-\exp{(-1 -
  \frac{\delta}{\tau_0})}) - 1 - \frac{\delta}{\tau_0}$.

The function $\mathbb{W}_{-1}(\cdot)$ is the lower real part of the
Lambert W function~\cite{corless1996lambertw}. We prove this
proposition in the supplementary material. During inference, we
perform the $\mathcal{O}(T)$ backward pass repeatedly. The existence
of a fixed point means that we can assume the stationary PGDS is in
its steady state and replace the backward pass with an
$\mathcal{O}(1)$ computation\footnote{Several software packages
  contain fast implementations of the Lambert W function.} of the
fixed point $\zeta^*$. To make this assumption, we must also assume
that $l_{\boldsymbol{\cdot} k}^{(T+1)} \sim
\textrm{Pois}(\zeta^{\star}\,\tau_0\,\theta_k^{(T)})$ instead of
$l_{\boldsymbol{\cdot} k}^{(T+1)} = 0$.
%The assumption that
%the PGDS is in its steady state can be made by assuming that
%$l_{\boldsymbol{\cdot} k}^{(T+1)} \sim
%\textrm{Pois}(\zeta^{\star}\,\tau_0\,\theta_k^{(T)})$ instead of
%assuming $l_{\boldsymbol{\cdot} k}^{(T+1)} = 0$, as in the non-steady
%state version.
We note that an analogous steady-state approximation
exists for the LDS and is routinely exploited to reduce
computation~\cite{rugh1996linear}.\looseness=-1

\textbf{Gibbs sampling algorithm:} Given $Y$ and the hyperparameters,
Gibbs sampling involves resampling each auxiliary variable or model
parameter from its conditional posterior. Our algorithm involves a
``backward filtering'' pass and a ``forward sampling'' pass, which
together form a ``backward filtering--forward sampling'' algorithm. We
use $- \setminus \Theta^{(\geq t)}$ to denote everything excluding
$\boldsymbol{\theta}^{(t)}, \ldots, \boldsymbol{\theta}^{(T)}$.

\textit{Sampling the auxiliary variables:} This step is the ``backward
filtering'' pass. For the stationary PGDS in its steady state, we
first compute $\zeta^*$ and draw $(l_{\boldsymbol{\cdot} k}^{(T+1)} \g
-) \sim \textrm{Pois}(\zeta^{\star}\,\tau_0\,\theta_k^{(T)})$. For the
other variants of the model, we set $l_{\boldsymbol{\cdot}
  k}^{(T+1)}=\zeta^{(T+1)}=0$. Then, working backward from $t=T,
\ldots, 2$, we draw\looseness=-1
\begin{align}
  \label{ref:l_.k}
(l_{k \boldsymbol{\cdot}}^{(t)} \g - \setminus\, \Theta^{(\geq t)})
&\sim
\textrm{CRT}(y_{\boldsymbol{\cdot} k}^{(t)} + l_{\boldsymbol{\cdot}
  k}^{(t+1)}, \tau_0 \textstyle{\sum}_{k_2=1}^K \pi_{k
  k_2}\,\theta_{k_2}^{(t-1)}) \textrm{ and\ \,}\\
  \label{ref:l_.k_all}
  (l_{k1}^{(t)}, \ldots, l_{kK}^{(t)} \g - \setminus\, \Theta^{(\geq t)})
&\sim \textrm{Mult}(l_{k \boldsymbol{\cdot}}^{(t)},
(\tfrac{\pi_{k1}\,\theta_{1}^{(t-1)}}{\sum_{k_2=1}^K \pi_{k
    k_2}\,\theta_{k_2}^{(t-1)}}, \ldots,
\tfrac{\pi_{kK}\,\theta_{K}^{(t-1)}}{\sum_{k_2=1}^K \pi_{k
    k_2}\,\theta_{k_2}^{(t-1)}})).
\end{align}
After using equations~\ref{ref:l_.k} and~\ref{ref:l_.k_all} for all
$k=1, \ldots, K$, we then set $l_{\boldsymbol{\cdot} k}^{(t)} =
\textstyle{\sum}_{k_1=1}^{K} l_{k_1 k}^{(t)}$. For the
non-steady-state variants, we also set $\zeta^{(t)} = \ln{(1 +
  \frac{\delta^{(t)}}{\tau_0} + \zeta^{(t+1)})}$; for the steady-state
variant, we set $\zeta^{(t)}=\zeta^*$.\looseness=-1
% \begin{equation}
%    \textrm{ and\ \,} \zeta^{(t)} = \ln{(1 + \frac{\delta^{(t)}}{\tau_0} + \zeta^{(t+1)})}.
%   \end{equation}

\textit{Sampling $\Theta$:} We sample $\Theta$ from its conditional
posterior by performing a ``forward sampling'' pass, starting with
$\boldsymbol{\theta}^{(1)}$. Conditioned on the values of
$l_{\boldsymbol{\cdot}k}^{(2)}, \ldots,
l_{\boldsymbol{\cdot}k}^{(T+1)}$ and $\zeta^{(2)}, \ldots,
\zeta^{(T+1)}$ obtained via the ``backward filtering'' pass, we sample
forward from $t=1, \ldots, T$, using the following
equations:\looseness=-1
\begin{align}
  \label{eqn:cond_post_theta_1}  
  (\theta_k^{(1)} \g - \setminus\, \Theta) &\sim
  \textrm{Gam}(%m_k^{(1)}
  y_{\boldsymbol{\cdot} k}^{(1)} + l_{\boldsymbol{\cdot} k}^{(2)}
  + \tau_0\,\nu_k, \tau_0
  + \delta^{(1)} + \zeta^{(2)}\,\tau_0)\textrm{ and\ \,}\\
  \label{eqn:cond_post_theta_t}
  (\theta_k^{(t)} \g - \setminus\, \Theta^{(\geq t)})
   &\sim \textrm{Gam}(%m_k^{(t)} +
  y_{\boldsymbol{\cdot}k}^{(t)} + l_{\boldsymbol{\cdot} k}^{(t+1)} + 
  \tau_0\textstyle{\sum}_{k_2=1}^K \pi_{k k_2}\,\theta_{k_2}^{(t-1)},
  \tau_0 + \delta^{(t)} + \zeta^{(t+1)}\,\tau_0).
\end{align}

\textit{Sampling $\Pi$:} The alternative model specification, with
$\Theta$ marginalized out, assumes that $(l_{1k}^{(t)}, \ldots,
l_{Kk}^{(t)}) \sim \textrm{Mult}(l_{\boldsymbol{\cdot} k}^{(t)},
(\pi_{1k}, \ldots, \pi_{Kk}))$. Therefore, via Dirichlet--multinomial
conjugacy,
\begin{equation}
\label{eqn:pi_k}
  (\boldsymbol{\pi}_{k} \g - \setminus\, \Theta) \sim \textrm{Dir}(\nu_1\nu_k +
  \textstyle{\sum}_{t=1}^T l_{1k}^{(t)}, \ldots, \xi\nu_k +
  \textstyle{\sum}_{t=1}^T l_{kk}^{(t)}, \ldots, \nu_K\nu_k +
  \textstyle{\sum}_{t=1}^T l_{Kk}^{(t)}).
\end{equation}

\textit{Sampling $\boldsymbol{\nu}$ and $\xi$:} We use the alternative
model specification to obtain closed-form conditional posteriors for
$\nu_k$ and $\xi$. First, we marginalize over $\boldsymbol{\pi}_k$ to
obtain a Dirichlet--multinomial distribution. When augmented with a
beta-distributed auxiliary variable, the Dirichlet--multinomial
distribution is proportional to the negative binomial
distribution~\cite{zhou2016nonparametric}. We draw such an auxiliary
variable, which we use, along with negative binomial augmentation
schemes, to derive closed-form conditional posteriors for $\nu_k$ and
$\xi$. We provide these posteriors, along with their derivations, in
the supplementary material.

We also provide the conditional posteriors for the remaining model
parameters---$\Phi$, $\delta^{(1)}, \ldots, \delta^{(T)}$, and
$\beta$---which we obtain via Dirichlet--multinomial, gamma--Poisson,
and gamma--gamma conjugacy.~\looseness=-1

\section{Experiments}
\label{sec:experiments}

In this section, we compare the predictive performance of the PGDS to
that of the LDS and that of gamma process dynamic Poisson factor
analysis (GP-DPFA)~\cite{acharya2015nonparametric}. GP-DPFA models a
single count in $Y$ as $y_v^{(t)} \sim \textrm{Pois}(\sum_{k=1}^K
\lambda_k\, \phi_{vk}\,\theta_k^{(t)})$, where each component's
time-step factors evolve as a simple gamma Markov chain, independently
of those belonging to the other components: $\theta_k^{(t)} \sim
\textrm{Gam}(\theta_k^{(t-1)}, c^{(t)})$. We consider the stationary
variants of all three models.\footnote{We used the \texttt{pykalman}
  Python library for the LDS and implemented GP-DPFA ourselves.} We
used five data sets, and tested each model on two time-series
prediction tasks: smoothing---i.e., predicting $y_v^{(t)}$ given
$y_v^{(1)}, \ldots, y_v^{(t-1)}, y_v^{(t+1)}, \ldots, y_v^{(T)}$---and
forecasting---i.e., predicting $y_v^{(T+s)}$ given $y_v^{(1)}, \ldots,
y_v^{(T)}$ for some $s \in \{1, 2, \ldots\}$~\cite{durbin2012time}.
We provide brief descriptions of the data sets below before reporting
results.\looseness=-1

\textit{Global Database of Events, Language, and Tone (GDELT):} GDELT
is an international relations data set consisting of
country-to-country interaction events of the form ``country $i$ took
action $a$ toward country $j$ at time $t$,'' extracted from news
corpora. We created five count matrices, one for each year from 2001
through 2005. We treated directed pairs of countries $i{\rightarrow}j$
as features and counted the number of events for each pair during each
day. We discarded all pairs with fewer than twenty-five total events,
leaving $T=365$, around $V \approx 9,000$, and three to six million
events for each matrix.

\textit{Integrated Crisis Early Warning System (ICEWS):} ICEWS is
another international relations event data set extracted from news
corpora. It is more highly curated than GDELT and contains fewer
events. We therefore treated undirected pairs of countries
$i{\leftrightarrow}j$ as features. We created three count matrices,
one for 2001--2003, one for 2004--2006, and one for 2007--2009. We
counted the number of events for each pair during each three-day time
step, and again discarded all pairs with fewer than twenty-five total
events, leaving $T=365$, around $V \approx 3,000$, and 1.3 to 1.5
million events for each matrix.\looseness=-1

\textit{State-of-the-Union transcripts (SOTU):} The SOTU corpus
contains the text of the annual SOTU speech transcripts from 1790
through 2014.  We created a single count matrix with one column per
year. After discarding stopwords, we were left with $T=225$,
$V=7,518$, and 656,949 tokens.\looseness=-1

\textit{DBLP conference abstracts (DBLP):} DBLP is a database of
computer science research papers. We used the subset of this corpus
that Acharya et al. used to evaluate
GP-DPFA~\cite{acharya2015nonparametric}. This subset corresponds to a
count matrix with $T=14$ columns, $V=1,771$ unique word types, and
13,431 tokens.\looseness=-1

\textit{NIPS corpus (NIPS):} The NIPS corpus contains the text of
every NIPS conference paper from 1987 to 2003. We created a single
count matrix with one column per year. We treated unique word types as
features and discarded all stopwords, leaving $T=17$, $V=9,836$, and
3.1 million tokens.

\begin{figure}[ht]
\centering
\includegraphics[width=\linewidth]{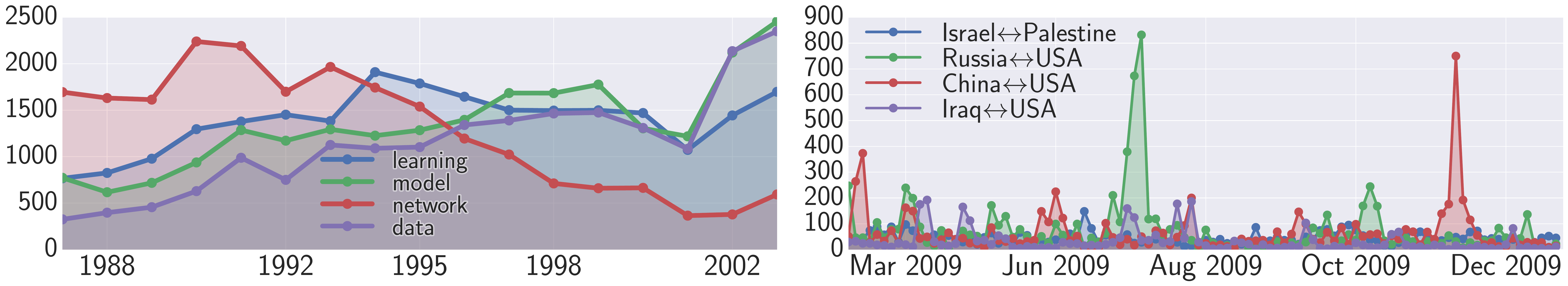}
\vspace{-4mm}
\caption{\label{fig:nips_vs_icews} $y_v^{(t)}$ over time for the top
  four features in the NIPS (left) and ICEWS (right) data
  sets.\looseness=-1}
\end{figure}

%\textbf{Polyphonic music (PM):} To test the models' predictive
%performance for binary observations, we used a polyphonic music
%sequence data set~\cite{boulanger2012modeling}. We created binary
%matrices for the first forty-four songs in this data set. Each matrix
%has one column per time step and one row for each key on the
%piano---i.e., $T=111$--$3,155$ and $V=88$. A single observation
%$b_v^{(t)} = 1$ means that key $v$ was played at time
%$t$.\looseness=-1

\textbf{Experimental design:} For each matrix, we created four masks
indicating some randomly selected subset of columns to treat as
held-out data. For the event count matrices, we held out six
(non-contiguous) time steps between $t=2$ and $t=T-3$ to test the
models' smoothing performance, as well as the last two time steps to
test their forecasting performance. The other matrices have fewer time
steps. For the SOTU matrix, we therefore held out five time steps
between $t=2$ and $t=T-2$, as well as $t=T$. For the NIPS and DBLP
matrices, which contain substantially fewer time steps than the SOTU
matrix, we held out three time steps between $t=2$ and $t=T-2$, as
well as $t=T$.

For each matrix, mask, and model combination, we ran inference four
times.\footnote{For the PGDS and GP-DPFA we used $K = 100$. For the
  PGDS, we set $\tau_0=1$, $\gamma_0 = 50$, $\eta_0 = \epsilon_0 =
  0.1$. We set the hyperparameters of GP-DPFA to the values used by
  Acharya et al.~\cite{acharya2015nonparametric}. For the LDS, we used
  the default hyperparameters for \texttt{pykalman}, and report
  results for the best-performing value of $K \in
  \{5,10,25,50\}$.\looseness=-1} For the PGDS and GP-DPFA, we
performed 6,000 Gibbs sampling iterations, imputing the missing counts
from the ``smoothing'' columns at the same time as sampling the model
parameters. We then discarded the first 4,000 samples and retained
every hundredth sample thereafter. We used each of these samples to
predict the missing counts from the ``forecasting'' columns. We then
averaged the predictions over the samples. For the LDS, we ran EM to
learn the model parameters. Then, given these parameter values, we
used the Kalman filter and smoother~\cite{kalman1960new} to predict
the held-out data. In practice, for all five data sets, $V$ was too
large for us to run inference for the LDS, which is $\mathcal{O}((K +
V)^3)$~\cite{ghahramani1999learning}, using all $V$ features. We
therefore report results from two independent sets of experiments: one
comparing all three models using only the top $V=1,000$ features for
each data set, and one comparing the PGDS to just GP-DPFA using all
the features. The first set of experiments is generous to the LDS
because the Poisson distribution is well approximated by the Gaussian
distribution when its mean is large.\looseness=-1

\setlength{\tabcolsep}{1.5pt}
\begin{table}[t]
   \caption{\label{fig:results} Results for the smoothing (``S'') and
     forecasting (``F'') tasks. For both error measures, lower values
     are better. We also report the number of time steps $T$ and the
     burstiness $\hat{B}$ of each data set.~\looseness=-1} \centering
   \footnotesize
  \begin{tabular}{rccccccccc}
    \toprule
    & & & & \multicolumn{3}{c}{Mean Relative Error (MRE)} & \multicolumn{3}{c}{Mean Absolute Error (MAE)}\\
        \cmidrule(lr){5-7} \cmidrule(lr){8-10}
     & $T$& $\hat{B}$ & Task & PGDS & GP-DPFA & LDS & PGDS & GP-DPFA & LDS \\
    \midrule
GDELT & 365 & 1.27 & S & $\mathbf{2.335}$ $\mathsmaller{\pm 0.19}$ & 2.951 $\mathsmaller{\pm 0.32}$ & 3.493 $\mathsmaller{\pm 0.53}$ & 9.366 $\mathsmaller{\pm 2.19}$ & $\mathbf{9.278}$ $\mathsmaller{\pm 2.01}$ & 10.098 $\mathsmaller{\pm 2.39}$ \\
 & & & F & $\mathbf{2.173}$ $\mathsmaller{\pm 0.41}$ & 2.207 $\mathsmaller{\pm 0.42}$ & 2.397 $\mathsmaller{\pm 0.29}$ & $\mathbf{7.002}$ $\mathsmaller{\pm 1.43}$ & 7.095 $\mathsmaller{\pm 1.67}$ & 7.047 $\mathsmaller{\pm 1.25}$ \\
\midrule
ICEWS & 365 & 1.10 & S  & $\mathbf{0.808}$ $\mathsmaller{\pm 0.11}$ & 0.877 $\mathsmaller{\pm 0.12}$ & 1.023 $\mathsmaller{\pm 0.15}$ & $\mathbf{2.867}$ $\mathsmaller{\pm 0.56}$ & 2.872 $\mathsmaller{\pm 0.56}$ & 3.104 $\mathsmaller{\pm 0.60}$\\
 & & & F & $\mathbf{0.743}$ $\mathsmaller{\pm 0.17}$ & 0.792 $\mathsmaller{\pm 0.17}$ & 0.937 $\mathsmaller{\pm 0.31}$ & $\mathbf{1.788}$ $\mathsmaller{\pm 0.47}$ & 1.894 $\mathsmaller{\pm 0.50}$ & 1.973 $\mathsmaller{\pm 0.62}$ \\
\midrule
SOTU & 225 & 1.45 & S & $\mathbf{0.233}$ $\mathsmaller{\pm 0.01}$ & 0.238 $\mathsmaller{\pm 0.01}$ & 0.260 $\mathsmaller{\pm 0.01}$ & $\mathbf{0.408}$ $\mathsmaller{\pm 0.01}$ & 0.414 $\mathsmaller{\pm 0.01}$ & 0.448 $\mathsmaller{\pm 0.00}$ \\
 & & & F & $\mathbf{0.171}$ $\mathsmaller{\pm 0.00}$ & 0.173 $\mathsmaller{\pm 0.00}$ & 0.225 $\mathsmaller{\pm 0.01}$ & 0.323 $\mathsmaller{\pm 0.00}$ & $\mathbf{0.314}$ $\mathsmaller{\pm 0.00}$ & 0.370 $\mathsmaller{\pm 0.00}$ \\
\midrule
DBLP & 14 & 1.64 & S & 0.417 $\mathsmaller{\pm 0.03}$ & 0.422 $\mathsmaller{\pm 0.05}$ & $\mathbf{0.405}$ $\mathsmaller{\pm 0.05}$ & $\mathbf{0.771}$ $\mathsmaller{\pm 0.03}$ & 0.782 $\mathsmaller{\pm 0.06}$ & 0.831 $\mathsmaller{\pm 0.01}$ \\
 & & & F & $\mathbf{0.322}$ $\mathsmaller{\pm 0.00}$ & 0.323 $\mathsmaller{\pm 0.00}$ & 0.369 $\mathsmaller{\pm 0.06}$ & 0.747 $\mathsmaller{\pm 0.01}$ & $\mathbf{0.715}$ $\mathsmaller{\pm 0.00}$ & 0.943 $\mathsmaller{\pm 0.07}$ \\
\midrule
NIPS & 17 & 0.33 & S & 0.415 $\mathsmaller{\pm 0.07}$ & $\mathbf{0.392}$ $\mathsmaller{\pm 0.07}$ & 1.609 $\mathsmaller{\pm 0.43}$ & 29.940 $\mathsmaller{\pm 2.95}$ & $\mathbf{28.138}$ $\mathsmaller{\pm 3.08}$ & 108.378 $\mathsmaller{\pm 15.44}$ \\
 & & & F & 0.343 $\mathsmaller{\pm 0.01}$ & $\mathbf{0.312}$ $\mathsmaller{\pm 0.00}$ & 0.642 $\mathsmaller{\pm 0.14}$ & 62.839 $\mathsmaller{\pm 0.37}$ & $\mathbf{52.963}$ $\mathsmaller{\pm 0.52}$ & 95.495 $\mathsmaller{\pm 10.52}$ \\
    \bottomrule
    \end{tabular}
  \end{table}

\textbf{Results:} We used two error measures---mean relative error
(MRE) and mean absolute error (MAE)---to compute the models' smoothing
and forecasting scores for each matrix and mask combination. We then
averaged these scores over the masks. For the data sets with multiple
matrices, we also averaged the scores over the matrices. The two error
measures differ as follows: MRE accommodates the scale of the data,
while MAE does not. This is because relative error---which we define
as $\tfrac{|y_{v}^{(t)}-\hat{y}_v^{(t)}|}{1 + y_{v}^{(t)}}$, where
$y_v^{(t)}$ is the true count and $\hat{y}_v^{(t)}$ is the
prediction---divides the absolute error by the true count and thus
penalizes overpredictions more harshly than underpredictions. MRE is
therefore an especially natural choice for data sets that are
bursty---i.e., data sets that exhibit short periods of activity that
far exceed their mean. Models that are robust to these kinds of
overdispersed temporal patterns are less likely to make
overpredictions following a burst, and are therefore rewarded
accordingly by MRE.\looseness=-1

In table~\ref{fig:results}, we report the MRE and MAE scores for the
experiments using the top $V=1,000$ features. We also report the
average burstiness of each data set. We define the burstiness of
feature $v$ in matrix $Y$ to be $\hat{B}_v =
\frac{1}{T-1} \sum_{t=1}^{T-1}
\frac{|y_v^{(t+1)}-y_{v}^{(t)}|}{\hat{\mu}_v}$, where $\hat{\mu}_v =
\frac{1}{T}\sum_{t=1}^T y_v^{(t)}$. For each data set, we calculated
the burstiness of each feature in each matrix, and then averaged these
values to obtain an average burstiness score $\hat{B}$. The PGDS
outperformed the LDS and GP-DPFA on seven of the ten prediction tasks
when we used MRE to measure the models' performance; when we used MAE,
the PGDS outperformed the other models on five of the tasks. In the
supplementary material, we also report the results for the experiments
comparing the PGDS to GP-DPFA using all the features. The superiority
of the PGDS over GP-DPFA is even more pronounced in these results. We
hypothesize that the difference between these models is related to the
burstiness of the data. For both error measures, the only data set for
which GP-DPFA outperformed the PGDS on both tasks was the NIPS data
set. This data set has a substantially lower average burstiness score
than the other data sets. We provide visual evidence in
figure~\ref{fig:nips_vs_icews}, where we display $y_v^{(t)}$ over time
for the top four features in the NIPS and ICEWS data sets. For the
former, the features evolve smoothly; for the latter, they exhibit
bursts of activity.\looseness=-1

\textbf{Exploratory analysis:} We also explored the latent structure
inferred by the PGDS. Because its parameters are positive, they
are easy to interpret. In figure~\ref{fig:nips}, we depict three
components inferred from the NIPS data set. By examining the time-step
factors and feature factors for these components, we see that they
capture the decline of research on neural networks between 1987 and
2003, as well as the rise of Bayesian methods in machine learning.
These patterns match our prior knowledge.\looseness=-1

\begin{figure}[ht]
\centering
\includegraphics[width=0.99\linewidth]{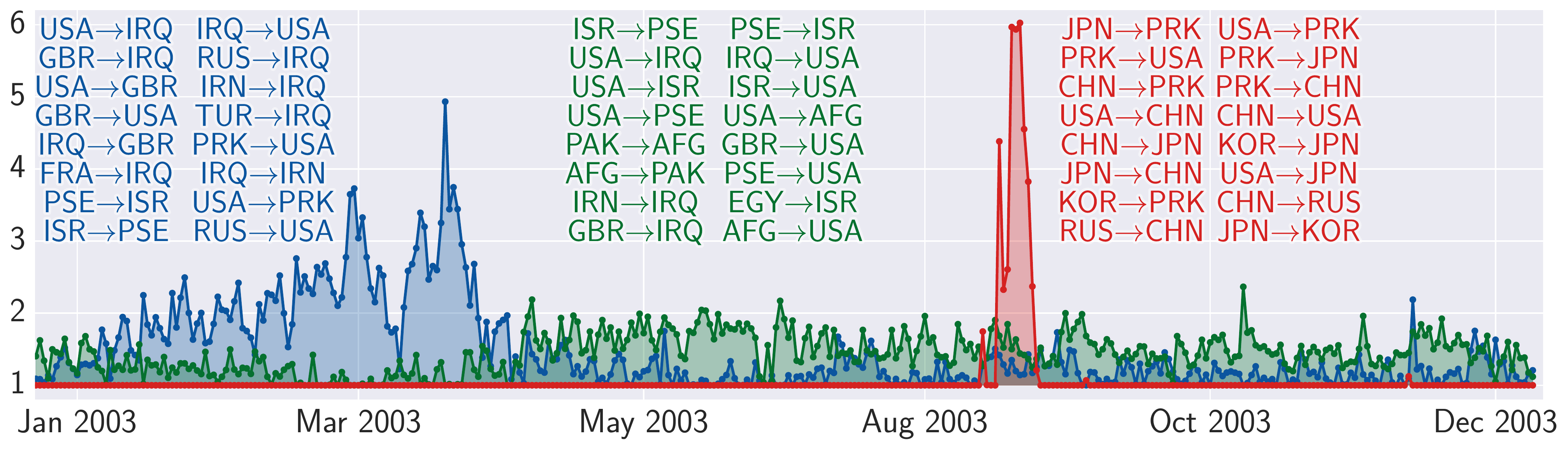}
\vspace{-1mm}
\caption{\label{fig:gdelt} The time-step factors for the top three
  components inferred by the PGDS from the 2003 GDELT matrix. The top
  component is in blue, the second is in green, and the third is in
  red. For each component, we also list the features (directed pairs
  of countries) with the largest feature factors.}
\end{figure}

In figure~\ref{fig:gdelt}, we depict the three components with the
largest component weights inferred by the PGDS from the 2003 GDELT
matrix. The top component is in blue, the second is in green, and the
third is in red. For each component, we also list the sixteen features
(directed pairs of countries) with the largest feature factors. The
top component (blue) is most active in March and April, 2003. Its
features involve USA, Iraq (IRQ), Great Britain (GBR), Turkey (TUR),
and Iran (IRN), among others. This component corresponds to the 2003
invasion of Iraq. The second component (green) exhibits a noticeable
increase in activity immediately after April, 2003. Its top features
involve Israel (ISR), Palestine (PSE), USA, and Afghanistan (AFG). The
third component exhibits a large burst of activity in August, 2003,
but is otherwise inactive. Its top features involve North Korea (PRK),
South Korea (KOR), Japan (JPN), China (CHN), Russia (RUS), and
USA. This component corresponds to the six-party talks---a series of
negotiations between these six countries for the purpose of
dismantling North Korea's nuclear program. The first round of talks
occurred during August 27--29, 2003.

\begin{wrapfigure}{r}{4cm}
\centering
\vspace{-2.9mm}
\includegraphics[width=\linewidth]{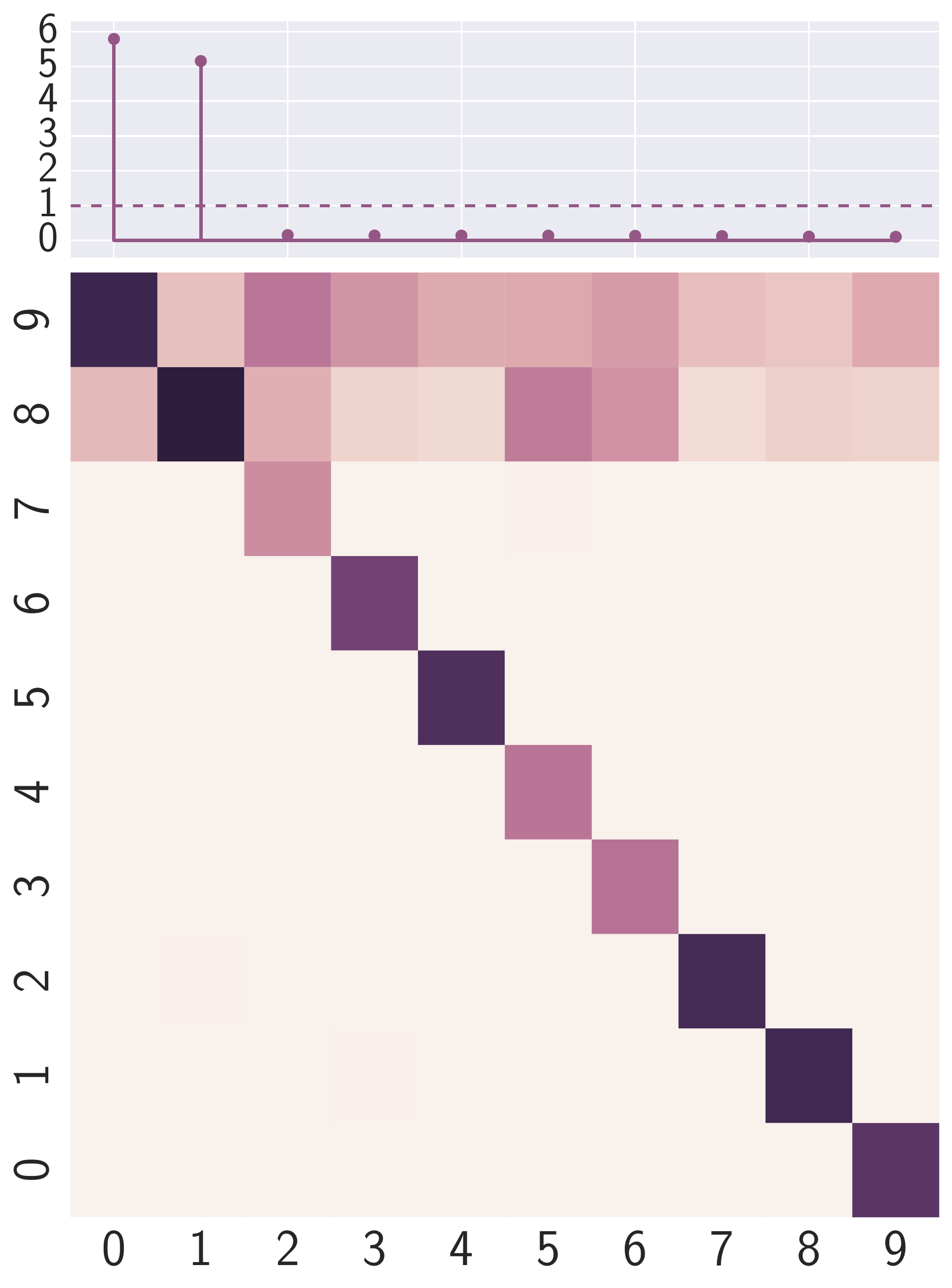}
\vspace{-4mm}
\caption{\label{fig:pi} The latent transition structure inferred by
  the PGDS from the 2003 GDELT matrix. Top: The component weights for
  the top ten components, in decreasing order from left to right; two
  of the weights are greater than one. Bottom: The transition weights
  in the corresponding subset of the transition matrix. This structure
  means that all components are likely to transition to the top two
  components.}
\vspace{-15mm}
\end{wrapfigure}
In figure~\ref{fig:pi}, we also show the component weights for the top
ten components, along with the corresponding subset of the transition
matrix $\Pi$. There are two components with weights greater than
one: the components that are depicted in blue and green in
figure~\ref{fig:gdelt}. The transition weights in the corresponding
rows of $\Pi$ are also large, meaning that other components are likely
to transition to them. As we mentioned previously, the GDELT data set
was extracted from news corpora. Therefore, patterns in the data
primarily reflect patterns in media coverage of international
affairs. We therefore interpret the latent structure inferred by the
PGDS in the following way: in 2003, the media briefly covered various
major events, including the six-party talks, before quickly returning
to a backdrop of the ongoing Iraq war and Israeli--Palestinian
relations. By inferring the kind of transition structure depicted in
figure~\ref{fig:pi}, the PGDS is able to model persistent, long-term
temporal patterns while accommodating the burstiness often
inherent to real-world count data. This ability is what enables the
PGDS to achieve superior predictive performance over the LDS and
GP-DPFA.\looseness=-1

\section{Summary}

We introduced the Poisson--gamma dynamical system (PGDS)---a new
Bayesian nonparametric model for sequentially observed multivariate
count data. This model supports the expressive transition structure of
the linear dynamical system, and naturally handles overdispersed
data. We presented a novel MCMC inference algorithm that remains
efficient for high-dimensional data sets, advancing recent work on
augmentation schemes for inference in negative binomial
models. Finally, we used the PGDS to analyze five real-world data
sets, demonstrating that it exhibits superior smoothing and
forecasting performance over two baseline models and infers highly
interpretable latent structure.\looseness=-1

\subsubsection*{Acknowledgments}

We thank David Belanger, Roy Adams, Kostis Gourgoulias, Ben Marlin, Dan Sheldon, and Tim Vieira for many helpful conversations.  This work was supported in part by the UMass Amherst CIIR and in part by NSF grants SBE-0965436 and IIS-1320219. Any opinions, findings, conclusions, or recommendations are those of the authors and do not necessarily reflect those of the sponsors.

\setlength{\bibsep}{4.4pt plus 0.3ex}
\begin{spacing}{-0.1}
\bibliographystyle{unsrt}
\bibliography{references}
\end{spacing}

\end{document}

%% file: macros.tex
\newcommand{\comment}[1]{}

\newcommand{\g}{\,|\,}
\newcommand{\teq}{\!=\!}

\newcommand{\Pois}{\textrm{Pois}}